\documentclass[letterpaper]{article} 
\usepackage{aaai25}  
\usepackage{times}  
\usepackage{helvet}  
\usepackage{courier}  
\usepackage[hyphens]{url}  
\usepackage{graphicx} 
\usepackage{natbib}  
\usepackage{caption} 
\usepackage{algorithm}
\usepackage{algorithmic}
\usepackage{amsmath}
\usepackage{amssymb}
\usepackage{booktabs}
\usepackage{tabularx}
\usepackage{enumitem}
\usepackage{multirow}
\usepackage{multicol}
\usepackage{array}
\usepackage{newfloat}
\usepackage{listings}
\DeclareCaptionStyle{ruled}{labelfont=normalfont,labelsep=colon,strut=off} 
\floatstyle{ruled}
\newfloat{listing}{tb}{lst}{}
\floatname{listing}{Listing}
\title{Bridge then Begin Anew: Generating Target-relevant Intermediate Model\\ for Source-free Visual Emotion Adaptation}
\author{
    Jiankun~Zhu\textsuperscript{\rm 1},
    Sicheng~Zhao\textsuperscript{\rm 2},
    Jing~Jiang\textsuperscript{\rm 1},
    Wenbo~Tang\textsuperscript{\rm 3},
    Zhaopan~Xu\textsuperscript{\rm 1},
    Tingting~Han\textsuperscript{\rm 4},
    Pengfei~Xu\textsuperscript{\rm 3},
    Hongxun~Yao\textsuperscript{\rm 1}\thanks{Corresponding author}
}
\affiliations{
    \textsuperscript{\rm 1}School of Computer Science and Technology, Harbin Institute of Technology, Harbin, China\\
    \textsuperscript{\rm 2}BNRist, Tsinghua University, Beijing, China\\
    \textsuperscript{\rm 3}NavInfo, Beijing, China\\
    \textsuperscript{\rm 4}Hangzhou Dianzi University, Hangzhou, China\\
    \{23s003059, 23s003045, 20b903054\}@stu.hit.edu.cn, h.yao@hit.edu.cn, schzhao@gmail.com, ttinghan@hdu.edu.cn, \{xupengfei11850, tangwenbo11883\}@navinfo.com
}

\usepackage{bibentry}

\begin{document}

\maketitle

\begin{abstract}
Visual emotion recognition (VER), which aims at understanding humans' emotional reactions toward different visual stimuli, has attracted increasing attention. Given the subjective and ambiguous characteristics of emotion, annotating a reliable large-scale dataset is hard. For reducing reliance on data labeling, domain adaptation offers an alternative solution by adapting models trained on labeled source data to unlabeled target data. Conventional domain adaptation methods require access to source data. However, due to privacy concerns, source emotional data may be inaccessible. To address this issue, we propose an unexplored task: source-free domain adaptation (SFDA) for VER, which does not have access to source data during the adaptation process. To achieve this, we propose a novel framework termed Bridge then Begin Anew (BBA), which consists of two steps: domain-bridged model generation (DMG) and target-related model adaptation (TMA). First, the DMG bridges cross-domain gaps by generating an intermediate model, avoiding direct alignment between two VER datasets with significant differences.
Then, the TMA begins training the target model anew to fit the target structure, avoiding the influence of source-specific knowledge. 
Extensive experiments are conducted on six SFDA settings for VER. The results demonstrate the effectiveness of BBA, which achieves remarkable performance gains compared with state-of-the-art SFDA methods and outperforms representative unsupervised domain adaptation approaches. 

\begin{links}
  \link{Code}{https://github.com/zhuzhu804/BBA}
\end{links}

\end{abstract}

%

\section{Introduction}
With the rapid development of social networks, people have become used to posting images to express their feelings~\cite{zhao2017real}. Visual emotion recognition (VER) aims to identify human emotions elicited by images~\cite{zhao2021affective}, attracting increasing attention and playing an essential role in various applications, such as depression detection~\cite{bokolo2023deep} and opinion mining~\cite{razali2021opinion}. Advanced technologies based on supervised deep neural networks have been proposed to improve the VER performance~\cite{deng2024learning,cen2024masanet}. Existing methods mainly follow a supervised pipeline requiring sufficient emotion annotations. However, labeling reliable large-scale datasets for VER tasks is challenging in practical applications because of the intrinsic properties of emotions, such as subjectivity, complexity, and ambiguity~\cite{zhao2023toward}. To reduce the annotation burden, much attention has been devoted to unsupervised domain adaptation (UDA) for VER, which enables the models trained on labeled source data to generalize well on the unlabeled target data~\cite{zhao2019cycleemotiongan}. However,
existing UDA methods require access to the source data during adaptation~\cite{li2023pseudo,zhao2024multi,jiang2024multi}. Regarding privacy and security concerns, source-free domain adaptation (SFDA) has attracted much interest in dealing with the situation where labeled source data is unavailable~\cite{li2024comprehensive}.

\begin{figure*}[!t]
\begin{center}
\centering \includegraphics[width=0.9\linewidth]{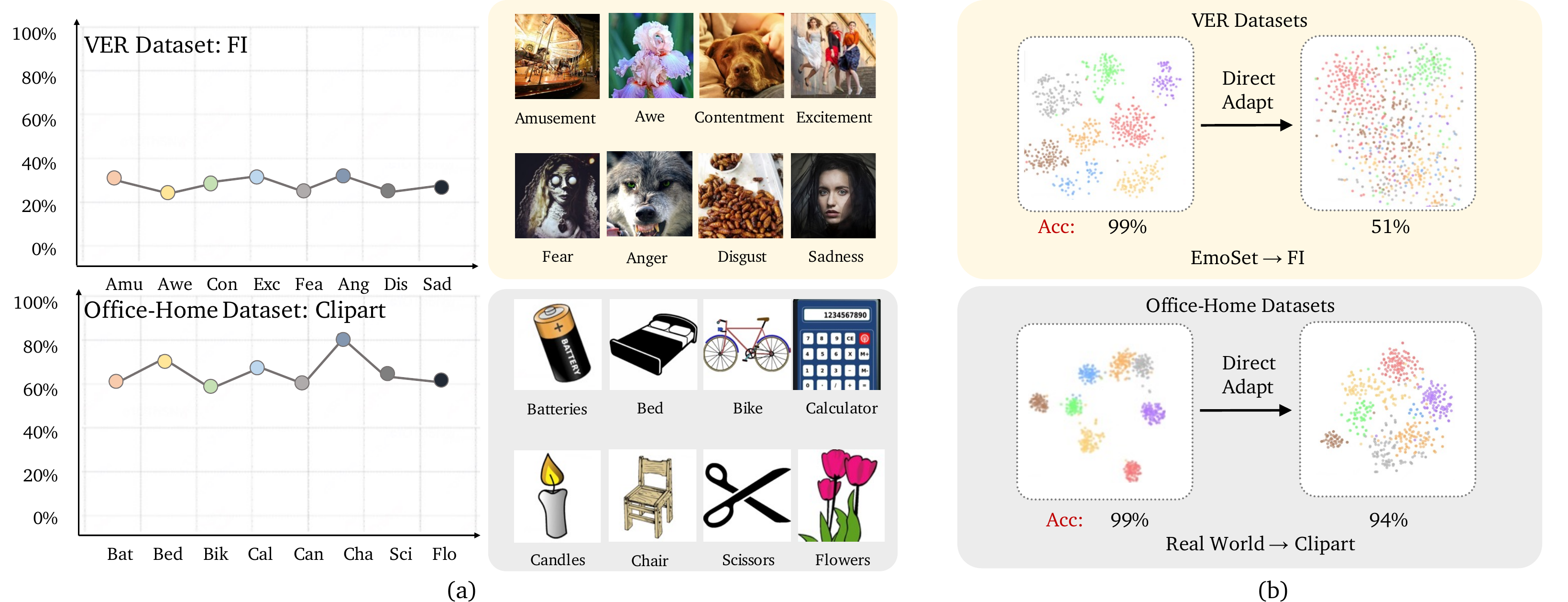}
\caption{Illustration of challenges for SFDA-VER tasks. (a) shows the pseudo-label confidence generated by SHOT on the VER dataset and the standard classification dataset OfficeHome. Differently colored dots indicate prediction probabilities for images in different categories.
(b) shows the distributions of ResNet-101 features on the VER and OfficeHome datasets. }
\label{fig:da}
\end{center}
\end{figure*}

On the one hand, although prior SFDA methods are effective for standard classification tasks, they face specific challenges and performance decreases when directly applied to VER. This is because of
the large affective gap~\cite{zhao2014exploring} between VER datasets, which arises from annotator variations and the scope of data collection. The distribution gap
significantly affects the feature alignment, which in turn reduces the confidence of pseudo-labels.
As shown in Figure~\ref{fig:da} (a), we use SHOT~\cite{liang2020we}, a typical SFDA method, to generate pseudo-labels for VER datasets EmoSet $\rightarrow$ FI and standard classification datasets Real World $\rightarrow$ Clipart, respectively. 
The results show that the classification scores for emotional pseudo-labels are relatively low, which increases the risk of misalignment between features and labels, and
can lead to a confirmation bias due to the accumulation of errors~\cite{ding2023combating}. To address this issue, bridging the inter-domain variation to generate reliable pseudo-labels is necessary.

On the other hand, traditional SFDA methods are fine-tuned on the source model. However, the unclear class features of the VER dataset make the source model lack robustness. 
For a better illustration, as shown in Figure~\ref{fig:da} (b), we train two ResNet-101 models~\cite{he2016deep} with the same number of categories on EmoSet and Clipart from the Office-Home~\cite{venkateswara2017deep}, respectively. The results indicate that the VER dataset has lower interclass distinction and higher intraclass variability. Consequently, the source model will overfit on noise, which is unrelated to the discriminative class features.
As a result, when directly adapted to the target dataset, \textit{i.e.,} EmoSet $\rightarrow$ FI and  Real World $\rightarrow$ Clipart, without further improvement, standard classification model tends to outperform VER model by a large margin. 
This suggests that the VER dataset lacks explicit class characterization, making the source domain model prone to overfitting.
To eliminate the effects of source domain noise, it is necessary for the target model to learn from the target data itself, instead of fine-tuning from the source model.

We summarize the challenges above: the large affective gap leads to incorrect pseudo-labels, and 
lack of clarity in the distribution of the VER dataset can lead to overfitting the source domain model.
To address these two challenges, we propose a novel SFDA framework for VER, termed Bridge then Begin Anew (BBA).
BBA contains two steps: 
Domain-bridged model generation (DMG) and target-related model adaptation (TMA).
To improve the confidence of pseudo-labels, DMG generates a bridge model to align emotional data across various domains, thus avoiding the challenges associated with directly fine-tuning emotional data. Moreover, we introduce a clustering-based
pseudo-label post-processing and masking strategy to constrain the pseudo-label distribution and explore richer semantic contexts.  
Then, to eliminate the effects of overfitting the source model, TMA starts training anew, 
which allows the target model to learn feature relations independently, thus extending the exploration of the target-specific features. Furthermore, 
in order to better learn about the emotional features,
we introduce polarity
constraints to enhance the target model’s discriminative ability for emotion categories. 

The main contributions of our work are outlined below:
\begin{itemize}
\item We propose a new task, \textit{i.e.}, source-free domain adaptation for visual emotion recognition (SFDA-VER), which focuses on the cross-domain transfer of VER without accessing source data during adaptation.
\item We propose a two-step framework, comprising DMG and TMA, to improve the reliability of pseudo-labels and eliminate the effects of source model overfitting.
\item We conduct experiments on six SFDA-VER settings, and the results show that our approach outperforms existing state-of-the-art methods by an average of +3.03.
\end{itemize}

\section{Related Works}

\subsection{Visual Emotion Recognition} 

Deep learning and CNNs have revolutionized VER~\citep{zhao2023toward}. The field evolved from global feature extraction~\cite{you2015robust, zhu2017dependency, yang2018retrieving, rao2020learning} to focusing on emotion-rich local regions and their relationships~\cite{yang2018visual,zhao2019pdanet,rao2019multi,yao2019attention,zhang2020object}. Recently, CLIP-based approaches have emerged~\cite{deng2022simple,deng2024learning,cen2024masanet}.
However, all these methods require large amounts of accurately labeled emotional data to train the network. 
Therefore, domain adaptation methods are needed to alleviate the requirement for a large number of annotations during the training process.

\begin{figure*}[!t]
\begin{center}
\centering \includegraphics[width=0.89\linewidth
]{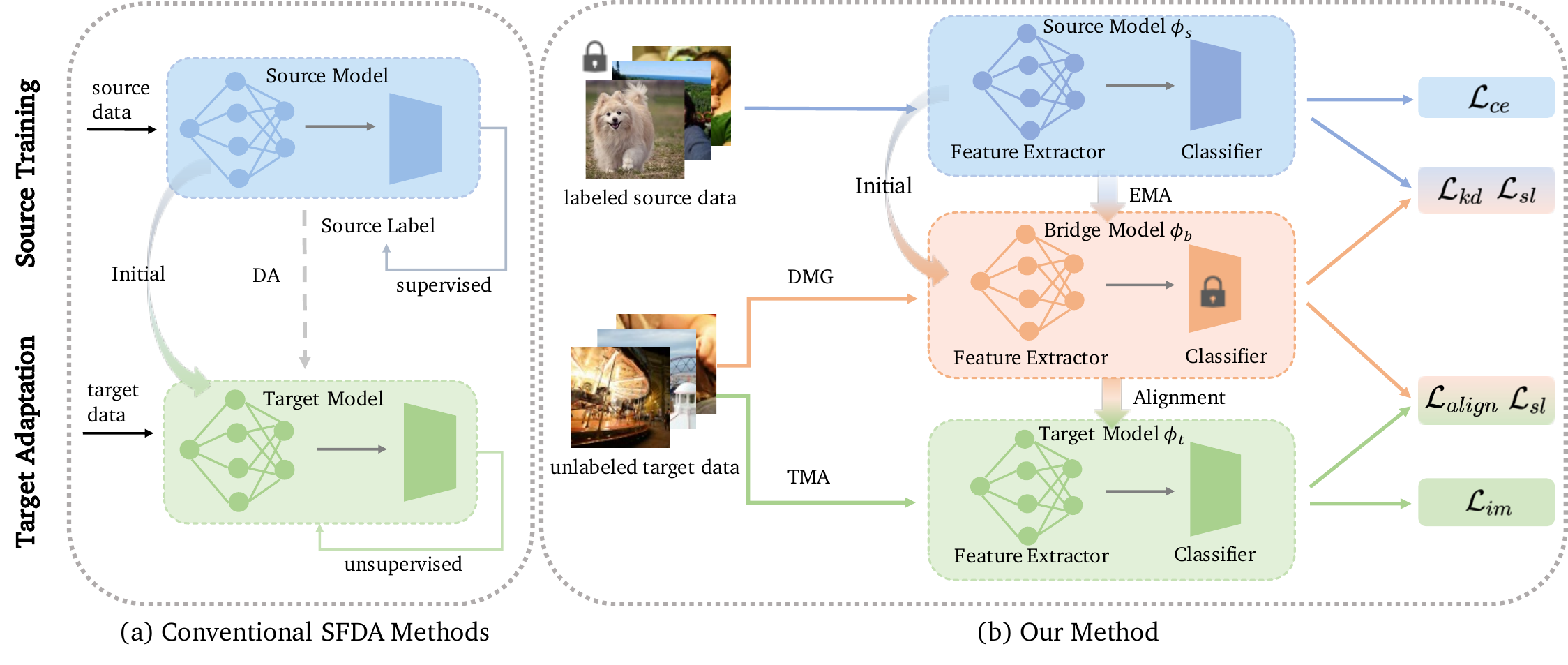}
\caption{
An overview comparison between conventional SFDA methods (a) and our method (b). In conventional methods (a), the source domain model is directly fine-tuned to align the source and target domains. This direct adaptation approach can be problematic due to significant differences between the source and target domains, potentially leading to suboptimal performance. In contrast, our Bridge then Begin Anew (BBA) approach (b) introduces a bridge model to generate more reliable pseudo-labels and stimulates the exploration of target domain-specific knowledge.}
\label{fig:Framework}
\end{center}
\end{figure*}

\subsection{Unsupervised Domain Adaptation for VER} 
Due to the significant domain bias in VER datasets, ~\citet{zhao2019cycleemotiongan} introduced CycleEmotionGAN, an unsupervised approach for cross-domain emotion adaptation using emotional consistency loss. ~\citet{zhao2021emotional} enhanced this with multi-scale similarity and improved emotional consistency. 
Considering the privacy constraints that limit direct access to source data for emotion analysis, our study presents a new source-free domain adaptation for VER tasks. It is required to adapt the source emotion recognition model to the target domain without accessing source data.

\subsection{Source Free Domain Adaptation} 
The SFDA setting represents a more complex but realistic UDA scenario where source data are unavailable~\cite{li2024comprehensive}. The following are some standard methods: Pseudo-labeling is to label the unlabeled target samples based on the predictions of the source model~\cite{liang2021source,huang2022relative,xie2022active,wang2022cross,ahmed2022cleaning,liang2022dine}. Entropy minimization is a constraint which can be inversely employed to guide the optimization of the model~\cite{liang2020we,jeon2021unsupervised,mao2024source,kothandaraman2021ss}. In virtual source methods~\cite{du2023generation,hao2021source}, the variants of Generative Adversarial Networks (GANs)~\cite{li2020model,kurmi2021domain} are common approaches to construct virtual source data. However, the above methods are insufficient to address the task of VER due to low domain relationships between emotional domains. Unlike existing work, we focus on mining the inherent structure of the target domain.

\section{Methods}

\subsection{Problem Definition} 
In this paper, we focus on Source-Free Domain Adaptation for Visual Emotion Recognition (SFDA-VER), adapting from a labeled source domain to an unlabeled target domain. 
Our source dataset contains ${N_S}$ labeled images $\left\{x_s^i\right\}_{i=1}^{N_S}$ with corresponding emotion labels $\left\{y_s^i\right\}_{i=1}^{N_S}$, where $x_s^i \in \mathcal{X}_s$, $y_s^i \in \mathcal{Y}_s$. The target dataset contains ${N_T}$ unlabeled images $T = \left\{x_t^i\right\}_{i=1}^{N_T}$, where $x_t^i \in \mathcal{X}_t$. The feature spaces $\mathcal{X}_s$ and $\mathcal{Y}_s$ differ across domains. For simplicity, the superscript $i$ is henceforth omitted. In particular, during the adaptation process, the data from the source domain is inaccessible; only the source model can be used for adaptation.

\subsection{Overview} Figure~\ref{fig:Framework} summarizes our framework, Bridge then Begin Anew (BBA), which consists of two steps: domain-bridged model generation (DMG) and target-related model adaptation (TMA).
Unlike most previous SFDA methods that utilize the source model to initialize the target model, DMG introduces a bridge model to enhance the consistency between domains and produce reliable pseudo-labels. To reduce source model overfitting due to category noise, TMA starts by training the target model anew, learning from the target structure itself independently.

\subsection{Domain-bridged Model Generation}\label{sec:2}
It is still a challenge for SFDA to generate high-confident pseudo-labels, especially for tasks with large inter-domain variations like VER. In traditional methods, the source model is directly used to generate pseudo-labels on the target data. These approaches make the confidence of pseudo-labels highly dependent on the generalization ability of the source model. Due to category noise in VER data, the source model lacks robustness, resulting in low-confidence pseudo-labels. Hence, we propose a bridge model to minimize domain gaps and generate high-confidence pseudo-labels.

In order to improve interclass distinction and reduce intraclass variability in VER features, we consider mitigating the pseudo-label mismatch problem caused by domain differences. Inspired by k-means, we propose a fused distance clustering method as a label optimization strategy, which imposes additional constraints on the model outputs. 
Specifically, we apply weighted k-means clustering to compute the centroid of each class within the target domain: 

\begin{equation}
c_k^{(0)} = \frac{\sum_{x_t \in X_t} \sigma_k(\phi_s(x_t)) g_s(x_t)}{\sum_{x_t \in X_t} \sigma_k(\phi_s(x_t))},
\end{equation}
where 
$
\sigma_k(a) = \frac{\exp(a_k)}{\sum_i \exp(a_i)}
$
denotes the \(k\)-th element in the softmax output of a \(K\)-dimensional vector \(a\), $\phi_{s} = f_s \circ g_s$ indicates source model, $g_s$ is the source feature extractor. $X_t = \{x_t^i\}^{N_T}_{i=1}$, $\hat{y}_t = \arg\max\sigma(\phi_{s}(x_t))$, $\sigma$ is the softmax operation, and $1$ is the indicator function.
After calculating the centroid of each class, we assign each sample to the nearest class as follows:
\begin{equation}
\hat{y}_t' = \arg\min_k D_f(g_{s}(x_t), c_k^{(0)}),
\end{equation}
where $D_f(P, Q)$ is the distance measure used to calculate the distance between each sample and centroid. Considering different distance metrics focus on various parts, we adopt a fused distance metric, \textit{i.e.}, we calculate Euclidean ($D_{eu}$), Cosine ($D_{cos}$), and Manhattan ($D_{man}$) distances between two features:
\begin{equation}
D_f(P, Q) = D_{eu}(P, Q) + D_{cos}(P, Q) + D_{man}(P, Q).
\end{equation}

Then, we update the category centroids according to:
\begin{equation}
c_k^{(1)} = \frac{\sum\limits_{x_t \in X_t} 1(\hat{y}_t' = k) g_{s}(x_t)}{\sum\limits_{x_t \in X_t} 1(\hat{y}_t' = k)}.
\end{equation}

\begin{figure}[t]
\includegraphics[width=\linewidth]{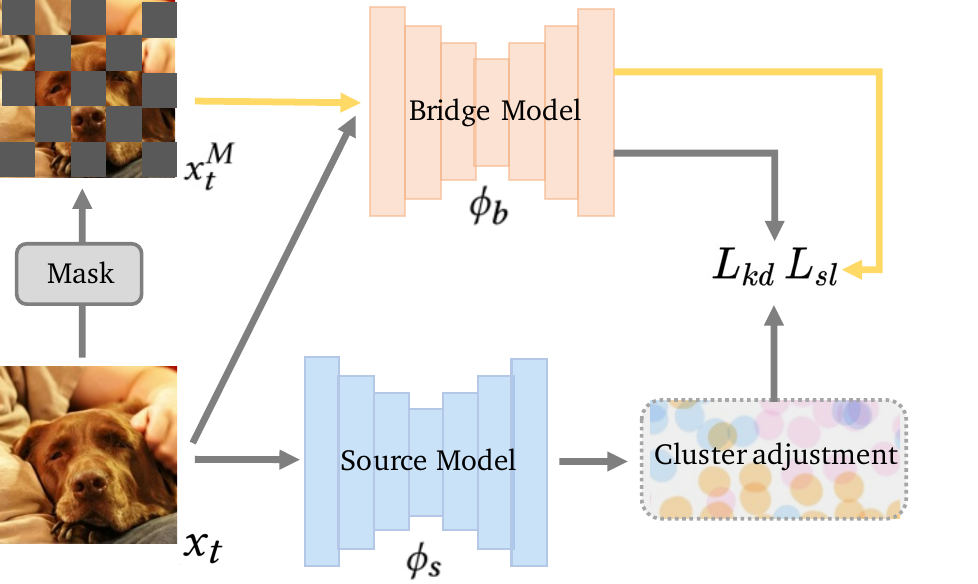}
  \caption{Illustration of our masking strategy for feature enhancement. The bridge model $\phi_b$, guided by the source model $\phi_s$, predicts both original and masked target domain images to compute self-labeling loss $\mathcal{L}_{sl}$ in Eq.~(\ref{eq:sl}) and distillation loss $\mathcal{L}_{kd}$ in Eq.~(\ref{eq:kd}), respectively.}
\label{fig:mask}
\end{figure}

To reduce intraclass variability, we consider the difference between VER images and traditional classification images: Unlike standard classification tasks, where different blocks in an image correspond to different semantics, different blocks in VER images are usually associated with a unified emotion label. As such, we propose a masking-enhanced framework to leverage emotional consistency across different contexts.
As shown in Figure~\ref{fig:mask}, models are encouraged to focus on local features to more effectively extract emotionally consistent features for better classification. Our method uses the random mask technique. Specifically, given a 2D image $I$ of size $H \times W$,
and the number of blocks to be masked $n$, which is randomly selected from a given range or distribution, the mask $M$ is defined as:
\begin{equation}M=\left\{m_{i j}\right\}_{i=1, j=1}^{H, W}, with\sum_{i=1}^{H}\sum_{j=1}^{W} m_{ij} = nb^2,\\\end{equation}
where $m_{i j} \in \left\{0,1\right\}$ is the mask value at position $(i, j)$, and $\text { if }(i, j) \in$ randomly selected blocks, $m_{i j} = 1$.
Each block has a size of $b \times b$. 
The masked image $x^M_t$ is obtained by the element-wise multiplication of $M$ and $x_t$, denoted as: 
\begin{equation}
x^M_t=M \odot x_t.
\end{equation}

After masking, the classifier is forced to use sparse residual cues,
to learn local and global affective information.
Moreover, to provide a consistent and stable guidance for bridge model, our source model adopts the exponential moving average (EMA) strategy for parameter updates~\cite{tarvainen2017mean}.
The bridge model $\phi_{b} = f_b \circ g_b$ inherits parameters from the pre-trained source model $\phi_{s}$. 
To better generalize to the target domain, the feature extractor $g_{b}$ is updated while the classifier $f_b$ is fixed. Both $g_{b}$ and $g_{s}$ are updated at this step, $g_{b}$ is updated by backpropagation as described in Eq.~(\ref{eq:loss_11}), and $g_{s}$ is updated by EMA: 
\begin{equation}
\theta_{t+1}^s \leftarrow \alpha_e\theta_t^s+(1-\alpha_e)\theta_{t}^b,
\end{equation}
where $t$ is the training step, $\alpha_e$ is the smoothing factor, $\theta_t^s$, $\theta_t^b$ are the parameters of $g_s$, $g_b$. 
Then, we train the masked image together with the original target domain image to participate in the computation of the self-labeling loss and the knowledge distillation loss:
\begin{equation}
\begin{split}
\mathcal{L}_{sl} = & -\mathbb{E}_{x_t \in X_t} \sum_{k=1}^K 1(\hat{y}_t'=k) \log \sigma(\phi_{b}(x_{t})) \\
& -\mathbb{E}_{x_t \in X_t} \sum_{k=1}^K 1(\hat{y}_t'=k) \log \sigma(\phi_{b}(x_t^M)),
\end{split}
\label{eq:sl}
\end{equation}
\begin{equation}
\mathcal{L}_{kd} = -\mathbb{E}_{x_t \in X_t} \hat{y}_t^{s} \log \phi_\text{b}(x_t)
-\mathbb{E}_{x_t \in X_t}  \hat{y}_t^{s} \log \phi_{b}(x^M_t),
\label{eq:kd}
\end{equation}
where $\hat{y}_t^{s} = D_f(g_{s}(x_t), c_k)$ denotes the soft pseudo output of the teacher feature extractor $g_s$ after clustering, and $c_k$ denotes the final centroid of each class.
The total loss function for the DMG is expressed as:
\begin{equation}
    \mathcal{L}_{dmg} = \mathcal{L}_{kd} + \lambda \mathcal{L}_{sl},
\label{eq:loss_11}
\end{equation}
where $\lambda$ is the weight for $\mathcal{L}_{sl}$, controlling the importance of $\mathcal{L}_{sl}$ during DMG step.

\subsection{Target-related Model Adaptation}\label{sec:3}
As stated earlier, noise in the source data that is unrelated to the discriminative features can cause the source model to overfit, potentially constraining its representation of target features.
To overcome it, we argue that the target model should be given sufficient latitude to explore the target feature space from scratch, enabling it to identify novel features. Following recent advances in self-distillation~\cite{kim2021self,laine2016temporal,liang2022dine}, we train the target model $\phi_t$ from scratch using target data alone, guided by the bridge model $\phi_b$, which produces more reliable pseudo-labels for improved supervision:
\begin{equation}
\mathcal L_{\mathrm{align}}=\mathbb{E}_{x_t \in X_t} D\left(\phi_b\left(x_t\right), \phi_t\left(x_t\right)\right),
\end{equation}
where $D(,)$ is a function that measures the distance between two emotion samples. In this paper, $D(,)$represents the KL divergence. $\phi_b$ updates pseudo-labels by mixing $\phi_t$ outputs, where $\alpha_t$ is a momentum hyper-parameter:
\begin{equation}
\phi_b\left(x_t\right) \leftarrow \alpha_t \phi_b\left(x_t\right)+(1-\alpha_t) \phi_t\left(x_t\right), \forall x_t \in X_t.
\end{equation}

\begin{figure}[!t]
\begin{center}
\centering \includegraphics[width=1.0\linewidth]{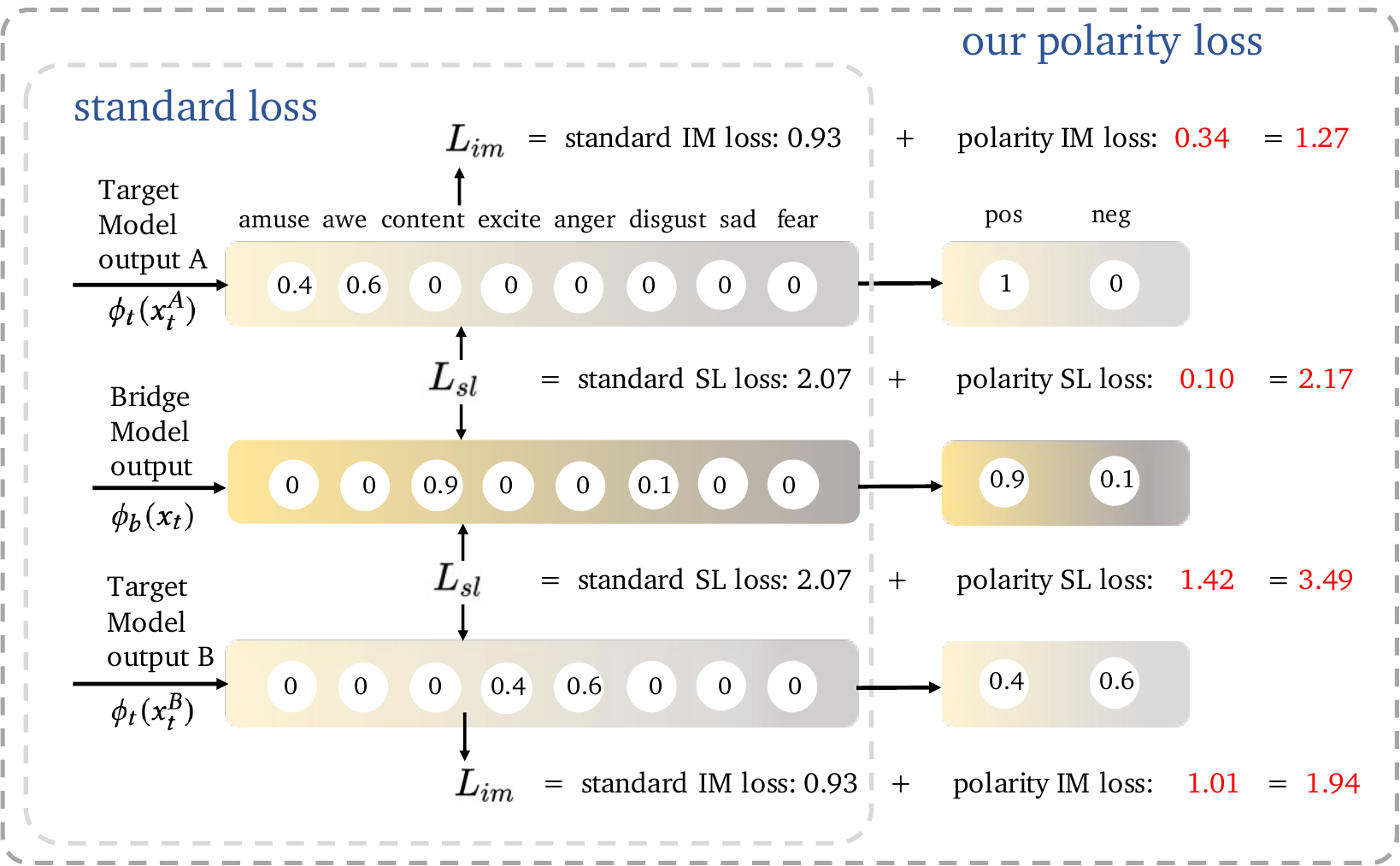}
\caption{Numerical examples to validate the effectiveness of the proposed emotion polarity loss. The addition of emotion polarity loss to IM loss $\mathcal{L}_{im}$ and SL loss $\mathcal{L}_{sl}$ allows the model to differentiate samples more finely.}
\label{fig:polarity_nn}
\end{center}
\end{figure}

In addition, we note that emotions naturally contain polarized information; categories with matching emotion polarities (positive or negative) show closer associations. To explore global and hierarchical polarity features when learning the target structure, we combine emotion polarity with the loss of self-labeling (SL) and the loss of information maximization (IM). Specifically, the distribution of emotion polarity is detailed as follows:
\begin{equation}
    p_{\text{pos}} = \sum_{k \in C_{\text{pos}}} \sigma(\phi_t(x_t))_k, 
    p_{\text{neg}} = \sum_{k \in C_{\text{neg}}} \sigma(\phi_t(x_t))_k,
\end{equation}
where \(C_{\text{pos}}\) and \(C_{\text{neg}}\) represents the positive and negative emotion categories. 
\(p_{pos}\) and \(p_{neg}\) symbolize the aggregate predicted likelihood across all positive and negative emotion categories for each sample, and they are collectively referred to $p_e$.
Then, we propose a polarized IM loss that incorporates the emotion polarity distribution: 
\begin{equation}
     \mathcal{L}_{im} = -\mathbb{E}_{x_t \in X_t} (\mathcal{L}_{e} + \mathcal{L}_{d}),
\end{equation}
\begin{equation}
     \mathcal{L}_{e} = \mathbb{E}_{k \in K} [p_k \log(p_k)]+\mathbb{E}_{e \in \{pos, neg\}} [p_e \log(p_e)],
\end{equation}
\begin{equation}
     \mathcal{L}_{d} = -\mathbb{E}_{k \in K} [p_k\log \mathbb{E}_k (p_k)]-\mathbb{E}_{e \in \{pos, neg\}} [p_e\log \mathbb{E}_e (p_e)],
\end{equation}
where $\mathcal{L}_{e}$ and $\mathcal{L}_{d}$ denote the entropy and diversity loss respectively. $p_k$ represents the predicted probability of the $k^{th}$ class, represented by $\sigma(\phi_t(x_t))_k$.
Moreover, we also incorporate the emotion polarity distribution into the SL loss, resulting in the following polarized SL loss:
\begin{equation}
    \mathcal{L}_{sl} =  -\mathbb{E}_{x_t \in X_t} \sum_{k=1}^K \hat{y}_t \log p_k-\mathbb{E}_{x_t \in X_t} \sum\limits_{e \in \{\text{pos},\text{neg}\}} \hat{y}_e \log p_e,
\end{equation}
where $\hat{y}_e = \arg\max p_e$ denotes the polarity pseudo-label. To illustrate the necessity and effectiveness of the proposed emotion polarity loss, we cite a simple example in Figure~\ref{fig:polarity_nn}.
We can see that the standard IM loss $\mathcal{L}_{im}$ gives the same loss value for different outputs $\phi_t(x_t^A)$ and $\phi_t(x_t^B)$, which are both 0.93. This is because the standard IM loss only focuses on the entropy of the output vector and overlooks the category relations. However, the polarity loss, which has values of 0.34 and 1.01, imposes an extra constraint on the model output. Furthermore, for the standard SL loss $\mathcal{L}_{sl}$, it is unchanged between the outputs $\phi_t(x_t^A)$, $\phi_b(x_t)$ and $\phi_t(x_t^B)$, $\phi_b(x_t)$, remaining at 2.07. After including the polarity loss, which are 0.10 and 1.42, the total loss differs. This highlights how the polarity loss considers an additional correlation across categories.

Overall, the total loss for the TMA phase can be expressed as:
\begin{equation}
\mathcal{L}_{tma} = \mathcal{L}_{align} + \mathcal{L}_{pol}, 
\label{eq:tma}
\end{equation}
where $\mathcal{L}_{pol} = \gamma \mathcal{L}_{sl} + \delta \mathcal{L}_{im}$, $\gamma$ and $\delta$ are the weights for $\mathcal{L}_{sl}$ and $\mathcal{L}_{im}$.

\subsection{BBA Learning}
For training, in the DMG step, we use clustered post-processed pseudo-labels to guide the training of both the target and masked target data. This approach aims to improve interclass separability and intraclass compactness, thereby generating more reliable pseudo-labels.
The loss function is $\mathcal{L}_{dmg}$ in Eq.~(\ref{eq:loss_11}). In the TMA step, to avoid the effects of source domain model overfitting, the target model is randomly initialized and guided by DMG-corrected pseudo-labels to learn the target structure.
The loss function is $\mathcal{L}_{tma}$ in Eq.~(\ref{eq:tma}). 
For inference, we only use the target model, and the model size is maintained the same as the source model.

\begin{table*}[!t]
\small

\caption{Performance comparison between BBA and state-of-the-art approaches on FI $\leftrightarrow$ EmoSet.}
\centering
\setlength{\tabcolsep}{1.8mm}
\begin{tabular}{l|c|c|ccc|ccc|c|ccc|ccc}
\hline
\multirow{3}{*}{Method} & \multirow{3}{*}{SF} & \multicolumn{7}{c|}{EmoSet $\rightarrow$ FI} & \multicolumn{7}{c}{FI $\rightarrow$ EmoSet} \\ \cline{3-16} 
                        &                     & \multirow{2}{*}{Acc} & \multicolumn{3}{c|}{macro avg} & \multicolumn{3}{c|}{weighted avg} & \multirow{2}{*}{Acc} & \multicolumn{3}{c|}{macro avg} & \multicolumn{3}{c}{weighted avg} \\ \cline{4-9} \cline{11-16}
                        &                     &                      & P & R & F1 & P & R & F1 & & P & R & F1 & P & R & F1 \\ \hline

Source only & {-} & 51.57 & 48.84 & 51.14 & 48.71 & 54.02 & 51.57 & 51.66 & 50.87 & 54.40 & 50.30 & 49.77 & 54.07 & 50.87 & 50.04 \\
Oracle & {-} & 67.32 & 62.71 & 59.44 & 60.55 & 66.50 & 67.32 & 66.55 & 71.96 & 72.60 & 72.40 & 72.41 & 72.12 & 71.96 & 71.95 \\
\hline
SHOT& \checkmark & 50.38 & 47.38 & 50.78 & 47.74 & 53.25 & 50.38 & 50.68 & 53.31 & 54.01 & 53.40 & 52.28 & 54.42 & 53.31 & 52.54 \\
SHOT++& \checkmark & 50.42 & 48.00 & 52.17 & 48.25 & 55.30 & 50.42 & 51.31 & 54.15 & 54.17 & 54.68 & 53.66 & 54.70 & 54.15 & 53.71 \\
G-SFDA & \checkmark & 50.47 & 47.95 & 52.09 & 48.22 & 55.22 & 50.47 & 51.36 & 54.40 & 55.36 & 54.29 & 53.66 & 55.32 & 54.40 & 53.81 \\
DaC  & \checkmark & 50.24 & 47.41 & 50.42 & 47.55 & 53.57 & 50.24 & 50.62 & 51.55 & 52.13 & 51.69 & 51.49 & 52.21 & 51.55 & 51.44 \\
AaD  & \checkmark & 49.90 & 48.01 & 51.96 & 47.74 & 55.71 & 49.90 & 50.88 & 52.68 & 52.54 & 53.46 & 52.37 & 53.33 & 52.68 & 52.40 \\
DINE & \checkmark & 55.25 & 51.87 & \underline{54.67} & 52.19 & 58.11 & 55.25 & 55.70 & \underline{56.36} & \underline{59.14} & 56.01 & 55.42 & \underline{59.04} & 56.36 & 55.66 \\
TPDS & \checkmark & 51.45 & 48.52 & 51.51 & 48.65 & 54.21 & 51.45 & 51.64 & 55.88 & 58.52 & 55.64 & 55.14 & 58.36 & 55.88 & 55.27 \\
BBA (ours) & \checkmark & \textbf{57.36} & \underline{52.69} & \textbf{55.67} & \textbf{53.27} & \textbf{59.92} & \textbf{57.36} & \textbf{57.76} & \textbf{58.76} & \textbf{60.32} & \textbf{58.83} & \textbf{58.52} & \textbf{59.62} & \textbf{58.76} & \textbf{58.19} \\
\hline
MCC  & {$\times$} & \underline{56.58} & 52.65 & 54.49 & 52.74 & 59.19 & \underline{56.58} & \underline{57.16} & 56.26 & 58.76 & 55.81 & 55.90 & 58.38 & 56.26 & 55.97 \\
ELS & {$\times$} & 56.42 & \textbf{53.82} & 53.94 & \underline{52.80} & 59.16 & 56.42 & 56.97 & 56.41 & 59.10 & \underline{56.50} & \underline{56.49} & 58.61 & \underline{56.41} & \underline{56.23} \\
MIC  & {$\times$} & 56.26 & 52.62 & 54.61 & 52.53 & \underline{59.85} & 56.26 & 57.41 & 54.31 & 54.26 & 54.77 & 53.74 & 54.73 & 54.31 & 53.78 \\
\hline

\end{tabular}
\label{Table 1}
\end{table*}

\begin{table*}[!t]
\centering
\small
\caption{Performance comparison between BBA and state-of-the-art approaches on FI $\leftrightarrow$ Emotion6.}
\setlength{\tabcolsep}{1.8mm}
\begin{tabular}{l|c|c|ccc|ccc|c|ccc|ccc}
\hline
\multirow{3}{*}{Method} & \multirow{3}{*}{SF} & \multicolumn{7}{c|}{Emotion6 $\rightarrow$ FI} & \multicolumn{7}{c}{FI $\rightarrow$ Emotion6} \\ \cline{3-16} 
                        &                     & \multirow{2}{*}{Acc} & \multicolumn{3}{c|}{macro avg} & \multicolumn{3}{c|}{weighted avg} & \multirow{2}{*}{Acc} & \multicolumn{3}{c|}{macro avg} & \multicolumn{3}{c}{weighted avg} \\ \cline{4-9} \cline{11-16}
                        &                     &                      & P & R & F1 & P & R & F1 & & P & R & F1 & P & R & F1 \\ \hline

Source only & {-} & 68.18 & 69.55 & 73.55 & 67.16 & 78.70 & 68.18 & 69.57 & 70.66 & 70.96 & 72.78 & 70.14 & 75.11 &  70.66 & 71.26\\
Oracle & {-} & 89.20 & 88.06 & 85.21 & 86.45 & 89.03 & 89.20 & 88.99 & 82.99 & 81.90 & 80.54 & 67.16 & 82.80 & 82.99 &82.81\\
\hline
SHOT & \checkmark & 67.22 & 68.97 & 72.80 & 66.26 & 78.24 & 67.22 & 68.63 & 67.90 & 70.15 & 71.52 & 67.72 & 74.93 & 67.90 & 68.41 \\
SHOT++ & \checkmark & 68.71 & 70.44 & 74.56 & 67.79 & 79.74 & 68.71 & 70.06 & 67.25 & 69.79 & 71.05 & 67.09 & 74.65 & 67.25 & 67.73 \\
G-SFDA & \checkmark & 65.39 & 68.64 & 72.10 & 64.71 & 78.37 & 65.39 & 66.75 & 66.89 & 69.62 & 70.81 & 66.75 & 74.53 & 66.89 & 67.35 \\
DaC& \checkmark & 64.43 & 67.66 & 70.94 & 63.73 & 77.33 & 64.43 & 65.82 & 65.93 & 69.80 & 70.58 & 65.88 & \underline{75.01} & 65.93 & 66.24 \\
AaD  & \checkmark & 70.04 & 71.04 & 75.39 & 68.98 & 80.06 & 70.04 & 71.36 & 67.84 & 69.92 & 71.33 & 67.64 & 74.65 & 67.84 & 68.37 \\
DINE& \checkmark & 71.71 & 71.31 & 75.76 & 70.30 & 79.79 & 71.71 & 72.99 & \underline{71.32} & \underline{71.28} & \underline{73.15} & \underline{70.72} & 74.97 & \underline{71.32} & \underline{71.91} \\
TPDS & \checkmark & \underline{72.42} & \underline{72.07} & \textbf{76.68} & 71.06 & \underline{80.55} & 72.42 & 73.67 & 63.23 & 69.04 & 68.92 & 63.23 & 74.65 & 63.23 & 63.18 \\
BBA (ours) & \checkmark & \textbf{78.12} & \textbf{73.84} & 75.64 & \textbf{74.55} & \textbf{80.59} & \textbf{78.12} & \textbf{78.52} & \textbf{73.23} & \textbf{72.49} & \textbf{74.38} & \textbf{72.43} & \textbf{76.15} & \textbf{73.23} & \textbf{73.77} \\
\hline
MCC  & {$\times$} & 74.04 & 70.54 & 73.70 & \underline{71.19} & 77.36 & \underline{74.04} & \underline{74.96} & 68.68 & 68.35 & 69.95 & 67.92 & 72.26 & 68.68 & 69.32 \\
ELS  & {$\times$} & 73.12 & 70.57 & 74.36 & 70.85 & 78.00 & 73.13 & 74.24 & 71.32 & 70.14 & 71.71 & 70.25 & 73.62 & \underline{71.32} & 71.85 \\
MIC & {$\times$} & 71.94 & 71.49 & \underline{75.97} & 70.52 & 79.94 & 71.94 & 73.21 & 65.21 & 68.56 & 69.46 & 65.13 & 73.58 & 65.21 & 65.59 \\
\hline
\end{tabular}
\label{Table 2}
\end{table*}


\section{Experiments}

\subsection{Experimental Settings}

Extensive experiments are conducted on four VER datasets: FI~\cite{you2016building}, EmoSet~\cite{yang2023emoset}, and ArtPhoto~\cite{machajdik2010affective} are categorized into Mikels' eight emotion categories~\cite{zhao2016predicting}; In contrast, Emotion6 is classified according to the Ekman model's six emotion categories~\cite{peng2015mixed}. We employ six SFDA-VER settings for experiments: FI $\leftrightarrow$ EmoSet, FI $\leftrightarrow$ ArtPhoto, and FI $\leftrightarrow$ Emotion6, with the latter considered as binary classification tasks due to their distinct categorical frameworks.

We compare BBA with the following baselines:
\textbf{1) Source only}, which refers to a basic approach where the model is trained on the source domain and directly applied to the target domain.
\textbf{2) SFDA methods}, which include SHOT~\cite{liang2020we}, SHOT++\cite{liang2021source}, G-SFDA\cite{yang2021exploiting}, AaD~\cite{yang2022attracting}, DaC~\cite{zhang2022divide}, DINE~\cite{liang2022dine}, and TPDS~\cite{tang2024source}.
\textbf{3) UDA methods}, which can utilize source domain data during the adaptation process, including MCC~\cite{jin2020minimum}, ELS~\cite{zhang2022free}, and MIC~\cite{hoyer2023mic}.
\textbf{4) Oracle}, which is the ideal scenario where the model is trained and tested within the target domain, representing the upper bound of performance.

In BBA, the EMA parameter $\alpha_e$ of the source network is set to 0.999. The loss weights $\lambda$, $\gamma$, and $\delta$ are set to 0.9, 1, and 0.3 for all six settings, respectively. More information about evaluation metrics and implementation details can be found in the supplementary material.

\begin{figure*}[!t]
\begin{center}
\centering 
\includegraphics[width=1.0\linewidth,trim=8.5cm 9.2cm 8.2cm 8.9cm, clip]{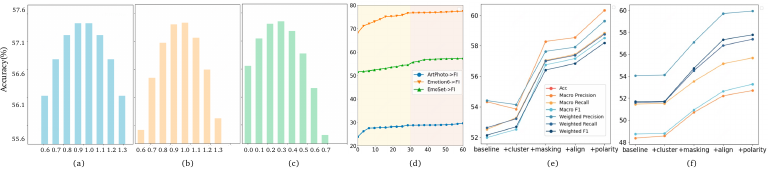}
\caption{Parameter analysis on EmoSet $\rightarrow$ FI (best viewed in color). (a): Analysis on $\lambda$. (b): Analysis on $\gamma$. (c): Analysis on $\delta$. (d): the accuracy during the training process. (e): Ablation study on EmoSet $\rightarrow$ FI. (f): Ablation study on FI $\rightarrow$ EmoSet.}
\label{fig:ab}
\end{center}
\end{figure*}

\begin{figure*}[!t]
\begin{center}
\centering \includegraphics[width=1.0\linewidth]{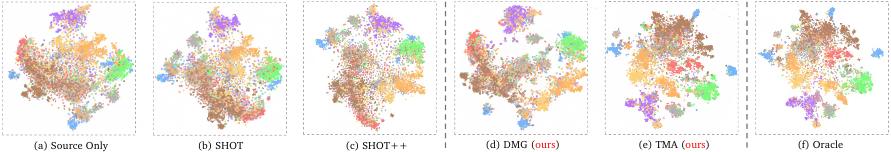}
\caption{t-SNE visualizations on FI $\rightarrow$ EmoSet. Different colors represent different classes.}

\label{fig:tsne_5}
\end{center}
\end{figure*}

\begin{table}[!t]
\footnotesize
\caption{Accuracy of different variants of our BBA on six SFDA settings. The methods are all added to the previous one. E, F, E6 means EmoSet, FI, and Emotion6 dataset, respectively.}

\begin{tabularx}{\columnwidth}{>{\hsize=1.2\hsize}X|c *{5}{>{\centering\arraybackslash\hsize=0.96\hsize}X}}

\hline
\multirow{2}{*}{Method} & \multicolumn{6}{c}{Accuracy(\%)} \\ \cline{2-7} 
                                             & E$\rightarrow$F & F$\rightarrow$E & A$\rightarrow$F & F$\rightarrow$A & E6$\rightarrow$F & F$\rightarrow$E6 \\
\hline

baseline & 51.59 & 52.63 & 27.40 & 33.95 & 66.81 & 70.42 \\
+cluster & 51.66 & 53.22 & 28.08 & 34.57 & 73.86 & 72.28 \\
+mask & 54.50 & 57.00 & 28.73 & 36.42 & 76.65 & 73.05 \\
+$\mathcal{L}_{align}$ & 56.79 & 57.38 & 28.91 & 37.04 & \textbf{78.12} & \textbf{73.23} \\
+$\mathcal{L}_{pol}$ & \textbf{57.36} & \textbf{58.76} & \textbf{29.63} & \textbf{37.65} & - & - \\
\hline
\end{tabularx}
\label{Table AB}
\end{table}
\vspace{-1mm}

\subsection{Comparisons with State-of-the-art Methods}

The results on FI $\leftrightarrow$ EmoSet and FI $\leftrightarrow$ Emotion6 are comprehensively detailed in Table~\ref{Table 1} and Table~\ref{Table 2}, respectively. More results for the other two settings, consisting of the small ArtPhoto dataset, can be found in the supplementary material.
Compared to the prior state-of-the-art SFDA methods, BBA achieves average improvements ranging from +2.26 to +7.17 on FI $\leftrightarrow$ EmoSet and from +3.81 to +11.3 on FI $\leftrightarrow$ Emotion6.
The increase is due to our 'bridge then begin anew' training framework, which improves the pseudo-labeling accuracy while learning from the target data distribution.
Despite the theoretical advantage of the UDA method in achieving better results by having access to the source domain data during the adaptation process, BBA outperforms the UDA method in accuracy, with average improvements of +1.57 to +2.78 on FI $\leftrightarrow$ EmoSet and +3.00 to +7.10 on FI $\leftrightarrow$ Emotion6, due to its targeted solutions to the challenges inherent in emotion data.

\subsection{Ablation Study}

In our ablation study, shown in Table \ref{Table AB}, we analyze the effectiveness of each component. We reveal that all modules are effective, with clustering and masking strategies bridging the domain gaps, enhancing feature discrimination, and generating more reliable pseudo-labels. Furthermore, $\mathcal{L}_{align}$ and $\mathcal{L}_{pol}$ enable the model to train directly on target domain data and explore the hierarchical features inherent in emotional images.
Panels (a), (b), and (c) of Figure~\ref{fig:ab} provide a sensitivity analysis for the loss weights $\lambda$, $\gamma$, and $\delta$. They show that the performance of BBA is relatively stable across different hyper-parameters. Panel (d) details the performance gains of the model during training, with different colors representing different stages of training. Panels (e) and (f) visualize the ablation results with different evaluation metrics on FI $\leftrightarrow$ EmoSet, further illustrating the effectiveness of the components proposed in BBA.

\subsection{Visualization}
\label{ssec:Visualization}

Figure~\ref{fig:tsne_5} visually analyzes the results before and after adaptation on FI $\rightarrow$ EmoSet. Panel (a) and Panel (f) illustrate the feature distribution on EmoSet for models trained on FI and EmoSet, respectively, showing the initial discrepancy between the source and target domains. 
Panels (b) and (c) show the feature distributions obtained by applying the SFDA methods SHOT and its extension SHOT++. Both methods retain substantial source domain knowledge due to their fundamental reliance on the source domain model.

Panel (d) visualizes the results after our DMG step. Compared with panels (b) and (c), DMG increases intra-class compactness and inter-class separability, making the feature distribution clearer and enabling the generation of high-confidence pseudo-labels. Panel (e) visualizes the results after our TMA step, highlighting its ability to effectively learn emotion feature representations from the target domain. This is achieved by learning the target domain model from scratch, closely mirroring the Oracle results shown in Panel (f). It demonstrates the superior ability of our method to capture the nuanced emotional features of the target domain, thereby setting a new state-of-the-art in SFDA-VER.

\section{Conclusion}
This paper introduces a new task termed source-free domain adaptation for visual emotion recognition (SFDA-VER). To address this task, we propose a novel method called BBA, which consists of two steps: domain-bridged model generation (DMG) and target-related model adaptation (TMA). First, the DMG step bridges the source and target domains. The clustering post-processing strategy enhances inter-class separability, while the masking strategy increases intra-class compactness. These improvements bolster the separability of categories and generate more reliable pseudo-labels. Subsequently, the TMA step focuses on the target structure to begin training anew. The emotion polarity loss enhances the model’s capabilities in emotion recognition. In summary, BBA addresses the unique challenges of the SFDA task within the VER dataset.

\section{Acknowledgments}
This research was supported by the National Science and Technology Major Project (2021ZD0110901) and the National Science Foundation of China (No. 62476069).

\bibliography{aaai25}

\end{document}